\documentclass[conference]{IEEEtran}
\IEEEoverridecommandlockouts
\usepackage{cite}
\usepackage{amsmath,amssymb,amsfonts}
\usepackage{algorithmic}
\usepackage{algorithm}
\usepackage{graphicx}
\usepackage{subcaption}
\usepackage{textcomp}
\usepackage{booktabs}
\usepackage{xcolor}
\usepackage{booktabs}
\usepackage{multirow}

\def\BibTeX{{\rm B\kern-.05em{\sc i\kern-.025em b}\kern-.08em
    T\kern-.1667em\lower.7ex\hbox{E}\kern-.125emX}}
\begin{document}

\title{PULSE: Personalized Opportunistic Learning for Wearable Devices}

\author{
    \IEEEauthorblockN{Mattia G. Campana\IEEEauthorrefmark{1}, Dariush Salami\IEEEauthorrefmark{2}, H\"{u}seyin Yi\u{g}itler\IEEEauthorrefmark{2}, Stephan Sigg \IEEEauthorrefmark{2}}
    \IEEEauthorblockA{
        \IEEEauthorrefmark{1}IIT-CNR, m.campana@iit.cnr.it\\ 
        \IEEEauthorrefmark{2}Aalto University, \{dariush.salami, huseyin.yigitler, stephan.sigg\}@aalto.fi\\
    }
}

\maketitle

\begin{abstract}
Wearable devices support personalized health and well-being applications, but post-deployment data can differ substantially from those used for training, while purely local adaptation may be limited by the data available on a single device during real-world operation. Conventional federated and decentralized learning rely on recurrent communication and stable connectivity, whereas existing opportunistic approaches often exploit available encounters without adequately considering their utility, compatibility, or communication cost.

We propose \emph{PULSE}, an event-driven framework for personalized opportunistic learning that treats each encounter as an online collaboration decision, selecting either no interaction or a promising nearby peer. Peer gradients are incorporated only when compatible with the local learning direction, while past interactions guide future peer selection. Across human activity recognition, fall detection, and stress recognition under synthetic and real contact traces, PULSE preserves or improves personalized classification performance, while reducing peer-assisted exchanges by 80--94\% compared with opportunistic model averaging. On-device experiments further demonstrate its feasibility on wearable hardware: peer-assisted operations require limited energy, while the complete implementation occupies 149.8~KiB of RAM and 561.8~KiB of flash. The obtained results show that selective opportunistic collaboration can provide effective personalization while limiting communication and resource demands on constrained wearable devices.
\end{abstract}


\section{Introduction}

Wearable devices support applications for personal well-being and health, such as activity recognition, physiological monitoring, and affective-state assessment, by continuously collecting sensor data~\cite{5306098, DBLP:journals/access/DemroziPBR20}. 
They are often designed as sensing and inference endpoints, with machine learning models trained before deployment and subsequently executed on-device to provide low-latency predictions while reducing the need to continuously transmit raw observations~\cite{10040604}. 
Yet, the data observed after deployment may differ substantially from those used during training because of person-specific behaviours and physiological characteristics, sensing hardware and placement, and changes in daily routines or operating conditions~\cite{ijcai2024p489, DBLP:journals/imwut/ChangMISK20}.

Although local adaptation can mitigate this issue, a single wearable may not collect sufficient labeled data to learn an effective personalized model in isolation~\cite{PRESOTTO2023101726}.
Collaborative learning offers a complementary source of information: devices can improve their models using knowledge acquired by others without directly sharing raw observations~\cite{11406160, 10.1561/2200000083}. 
%% I felt that the below is not necessary here and indeed distracts from the main line of discussion
%Federated learning (FL) provides a general framework for collaborative training across distributed clients~\cite{10.1561/2200000083}, while Personalized Federated Learning further recognizes that the optimal model may differ across users or data domains~\cite{11406160}. 
%However, conventional Federated Learning typically relies on a coordinating server, recurrent communication rounds, and reliable connectivity for model aggregation, assumptions that are difficult to reconcile with the energy, memory, and connectivity constraints of wearable devices.

%While wearables can often access the Internet through a smartphone or wireless infrastructure, cloud-mediated collaboration is not always desirable or available. It requires access to a trusted coordinator and can introduce wide-area communication, latency, privacy, and deployment costs that are unsuitable for applications expected to operate autonomously in everyday environments. Direct device-to-device communication provides an infrastructure-independent alternative, enabling connected peers to collaborate without relying on an external service.
%Decentralized Learning removes the central coordinator, but most existing approaches still assume persistent communication graphs or repeated gossip exchanges~\cite{10542323}, which remain impractical when contacts are intermittent and communication must be used sparingly.

Human mobility naturally brings nearby devices into contact, creating intermittent opportunities to exchange learning information and to support collaborative adaptation.
This setting has motivated \emph{Opportunistic Learning} (OL), in which decentralized model adaptation is driven by transient device-to-device encounters rather than scheduled communication rounds or persistent network connections~\cite{9439130}. Existing OL approaches have explored gradient exchange, contact-aware collaboration policies, and the transfer of model components among encountered devices~\cite{DBLP:conf/percom/RomeroM0SY25}.
While they demonstrate that intermittent encounters can support collaborative learning without centralized orchestration, the heterogeneity of wearable data makes indiscriminate knowledge transfer particularly problematic: a peer update may be irrelevant or harmful when different sensing conditions, activity distributions, or locally adapted representations are merged.

In this work we introduce \emph{PULSE}, a personalized opportunistic learning mechanism that makes peer collaboration explicitly selective at two levels. First, at each encounter, a lightweight utility-guided policy selects one promising peer or chooses no contact when communication is not expected to be useful. Second, rather than averaging models or accepting a peer update unconditionally, PULSE evaluates the agreement between the remote and local gradients and incorporates the remote signal only when it is positively aligned with the local learning direction. Each device fine-tunes its full model on private data, while peer interactions exchange only the current prediction head and its corresponding gradient, keeping the collaboration lightweight and preserving raw-data locality.

The contributions of this work are 

\begin{itemize}
    \item \textbf{PULSE}, an event-driven framework for Personalized OL that combines local adaptation with peer-assisted updates;

    \item \textbf{Selective collaboration} through a two-level mechanism that determines whether a peer should be contacted and whether its returned gradient should influence the local training direction;

    \item \textbf{Comprehensive simulation results} across three heterogeneous wearable applications and synthetic and real contact traces, considering both classification performance and communication demand;

    \item a \textbf{Prototype implementation} to characterize the latency, memory, and energy requirements of real wearables;

    \item an \textbf{Open Codebase}, including the full implementation and experimental code, publicly available at \emph{https://anonymous.4open.science/r/PULSE-F68A}.
\end{itemize}

\section{Related Work}
%\subsection{Federated and Decentralized Learning at the edge}
Federated Learning enables multiple devices to improve a learning model without directly sharing their raw data, typically by periodically communicating local model updates to a coordinating server~\cite{DBLP:conf/aistats/McMahanMRHA17}.
This paradigm can reduce the exposure of personal sensing data and has motivated extensive work on mobile and edge intelligence~\cite{10.1561/2200000083}.
In many contemporary edge deployments, pervasive connectivity and access to a coordinating infrastructure may indeed be available.
However, conventional Federated Learning typically relies on frequent communication rounds and on the transfer of model parameters or gradients, which can impose substantial communication, energy, and computation overhead.
Such requirements may be poorly aligned with pervasive sensing scenarios involving resource-constrained mobile and wearable devices, for which continuous participation in collaborative learning may be impractical or inefficient.

Decentralized Federated Learning removes the central coordinator and enables devices to exchange learning updates directly with peers~\cite{10542323}. This improves robustness to infrastructure failures and avoids reliance on a single aggregation point.
However, most decentralized optimization methods assume frequent peer-to-peer exchanges over persistent or repeatedly available communication graphs, typically through iterative gossip protocols.
While these assumptions may be reasonable for smartphones or single-board computers connected through Wi-Fi or local networks, repeated computation and communication can impose substantial energy costs on resource-constrained wearable devices~\cite{DBLP:journals/tmc/MishraG25}.

%\subsection{Opportunistic learning under transient contacts}
A line of work closer to our research considers collaboration driven by physical encounters among mobile devices. Lee \emph{et al.} introduced Opportunistic Federated Learning, in which devices personalize locally trained models by incorporating knowledge from encountered peers and use mobility and contact prediction to decide whether collaboration is worthwhile~\cite{9439130}.
This formulation importantly recognizes that encounters differ in their expected utility and duration, rather than treating communication opportunities as uniformly available. At the same time, the approach is centered on exchanging learning updates (i.e., gradients) between devices and relies on estimating future contact properties, which may be inaccurate or unavailable in highly dynamic environments.

More recently, OPTIMIST considers opportunistic federated transfer learning under intermittent connectivity and without centralized orchestration during the learning process~\cite{DBLP:conf/percom/RomeroM0SY25}.
It combines a frozen, pre-trained backbone with a trainable classifier head that is partitioned into non-overlapping subnetworks.
Each subnetwork is assigned to a node, trained locally, and transferred to an encountered peer, thereby enabling its sequential exposure to multiple local datasets while avoiding parameter conflicts among parallel updates. This design substantially reduces the computational burden relative to end-to-end collaborative training and is evaluated through both simulations and an Android-based prototype.

However, the framework presupposes an initialization procedure that creates the partition of the classifier head and assigns a distinct subnetwork to each participating node.
Although this procedure need not entail centralized coordination during subsequent encounters, it implicitly requires either a bootstrap component with global knowledge of the participating nodes or a shared deterministic assignment mechanism; the paper does not further specify how this initialization would be realized in a fully autonomous deployment. Moreover, OPTIMIST formulates learning as the optimization of a classifier head over the virtually merged data of all nodes.
Its primary objective is therefore collaborative generalization: subnetworks circulate to incorporate information from multiple datasets, rather than being selected or adapted according to the specific needs, data distribution, or compatibility of an individual user.
While nodes may hold different local head instances at a given time because of their distinct encounter histories, this divergence is an effect of asynchronous exchanges rather than an explicit personalization mechanism.
Finally, its mobile evaluation targets Android smartphones and parameter-level subnetwork transfer, whose resource requirements remain different from those of microcontroller-class wearable devices.

%\subsection{Prototype-based and communication-efficient knowledge sharing}
%A complementary stream reduces communication by replacing full-model updates with compact semantic representations. Prototype-based Federated Learning methods communicate class-level feature centroids, enabling clients to align local representations while accommodating statistical and, in some cases, architectural heterogeneity~\cite{DBLP:conf/aaai/TanLLZ00Z22}.
%Their main advantage is that the transmitted information can be much smaller than a neural-network update and may remain useful even when participants maintain personalized models.
%Recent decentralized approaches have explored prototype exchange and communication-reduction mechanisms within gossip- or neighbor-based learning schemes~\cite{10.1016/j.asoc.2025.113270, math13020237}.

%These approaches provide promising mechanisms for communication-efficient collaboration, but they are generally studied under recurring aggregation or gossip schedules and do not necessarily model volatile encounter processes. Moreover, the utility of a class-level representation depends on the quality and coverage of the local data: prototypes may be noisy for under-represented classes and may reflect user-specific sensing conditions under strong data heterogeneity. Thus, prototype exchange alleviates the payload bottleneck but does not by itself resolve which information should be exchanged, when it remains useful, or how it can be transferred within a short device encounter.

\subsection{Real-device evaluation of collaborative edge learning}
Experimental validation on physical devices is necessary to complement simulation-based analyses, which cannot fully capture the execution time, memory footprint, wireless latency, and energy consumption associated with collaborative learning. Prior work has developed Federated Learning testbeds on resource-constrained edge platforms, offering valuable insights into the discrepancy between abstract communication rounds and their practical realization~\cite{9829814, DBLP:journals/percom/TrottaMCBBF24}.
These studies enable an end-to-end assessment of system-level behavior beyond predictive performance alone. However, the considered platforms generally rely on single-board computers, smartphones, or other comparatively capable edge devices, typically operating over Wi-Fi or an accessible local network.

On the other hand, wearable sensing introduces a distinct operating regime: devices have tighter energy and memory budgets, may collect highly personal and user-dependent data, and encounter peers through short-lived, irregular contacts~\cite{DBLP:journals/iotj/NiATEN24}.
Consequently, realistic assessment should jointly account for learning quality, heterogeneous sensing data, contact dynamics, transmitted payload, execution latency, memory consumption, and energy expenditure.
This motivates evaluations that combine controlled simulations, including real-world mobility patterns or contact traces, with measurements on actual resource-constrained wearable hardware.

\section{The PULSE algorithm}

\begin{algorithm}[t]
\caption{PULSE at node $i$}
\label{algo:pulse}
\begin{algorithmic}[1]
\REQUIRE Local model $f_i=h_i\circ\mathcal{E}_i$, local data $\mathcal{D}_i$,
local training steps $\mathcal{R}$, peer-utility estimates $\mathcal{U}_i$

\item[]
\STATE \textbf{Upon} \textsc{OnLocalTraining}:
\STATE \hspace{1em}\textbf{for} $r = 1,\ldots,\mathcal{R}$ \textbf{do}
\STATE \hspace{2em}$f_i^{r+1} \gets$ Update $f_i^r$ with Eq.~\ref{eq:pulse-local-update}
\STATE \hspace{1em}\textbf{end for}

\item[]
\STATE \textbf{Upon} \textsc{OnPeerDiscovery}$(\mathcal{N}_i(t))$:
\STATE \hspace{1em}$j \gets \textsc{SelectPeer}(\mathcal{N}_i(t),\mathcal{U}_i)$
\STATE \hspace{1em}$\mathcal{B}_i \leftarrow$ Sample a minibatch from $\mathcal{D}_i$
\STATE \hspace{1em}$g_i \gets \nabla_h
\ell_i(h_i\circ\mathcal{E}_i;\mathcal{B}_i)$
\STATE \hspace{1em}\textbf{if} $j = \varnothing$ \textbf{then}
    \STATE \hspace{2em}$h_i \gets h_i - \eta g_i$
\STATE \hspace{1em}\textbf{else}
    \STATE \hspace{2em}\texttt{send} a copy of $h_i$ to $j$
    \STATE \hspace{2em}$g_j \gets$ \texttt{receive} remote gradient from $j$
    \STATE \hspace{2em}$a_{ij} \gets$ Gradient agreement with Eq.~\ref{eq:pulse-agreement}
    \STATE \hspace{2em}$\alpha_{ij} \gets$ Peer contribution with Eq.~\ref{eq:pulse-weight}
    \STATE \hspace{2em}$h_i \gets$ Update $h_i$ with Eq.~\ref{eq:pulse-mixing}
    \STATE \hspace{2em}$\mathcal{U}_i[j] \gets$ Update utility with Eq.~\ref{eq:pulse-utility-update}
\STATE \hspace{1em}\textbf{end if}
\end{algorithmic}
\end{algorithm}

PULSE is an event-driven mechanism specifically designed for personalized learning on wearable and edge devices experiencing brief and sporadic peer encounters.
Before deployment, a shared encoder is pretrained on representative data in a centralized bootstrap phase. Each node~$i$ then initializes its local model as $f_i=h_i\circ\mathcal{E}_i$, where $\mathcal{E}_i:\mathcal{X}\rightarrow\mathbb{R}^d$ is initialized from the pretrained encoder and maps an input observation to a $d$-dimensional latent representation, while a lightweight multilayer-perceptron head $h_i:\mathbb{R}^d\rightarrow\mathcal{Y}$ maps this representation to the output space.

After deployment, each node locally fine-tunes $f_i$ on its private, potentially evolving data distribution, allowing the model to adapt to user-specific behaviours and sensing conditions while retaining the task-aware initialization obtained during bootstrap. Rather than averaging complete models or exchanging raw observations, PULSE restricts peer-assisted interactions to gradients of $h_i$, thus keeping both the communication payload and peer-side computation lightweight.

Since information received from an arbitrary peer may be irrelevant or harmful, PULSE selectively incorporates remote learning signals. During an encounter, node~$i$ sends its current $h_i$ to a selected peer $j$, which evaluates a temporary copy using its local $\mathcal{E}_j$ and a private minibatch; node~$i$ then compares the resulting gradient with its local learning direction and incorporates it only when their cosine agreement is positive.

Algorithm~\ref{algo:pulse} summarizes the operations performed by node~$i$ upon local-training and peer-discovery events.

\subsection{Local Update}
At scheduled times, node~$i$ fine-tunes $f_i$ using its local dataset $\mathcal{D}_i$, thus supporting continuous adaptation even when useful peer encounters are unavailable.
At each local update event, node~$i$ performs $R$ stochastic-gradient steps, as detailed in lines~2--5 of Algorithm~\ref{algo:pulse}:

\begin{equation}
    f_i^{r+1} =
    f_i^r -
    \eta \nabla_{f}\ell_i(f_i^r;\mathcal{B}_i^r),
\label{eq:pulse-local-update}
\end{equation}

where $\eta$ is the learning rate, $\mathcal{B}_i^r$ is a minibatch sampled from $\mathcal{D}_i$, and $\ell_i$ denotes the local prediction loss, e.g., cross-entropy for classification.

\subsection{Opportunistic Peer-Assisted Update}

Whenever nearby devices are detected, node~$i$ forms the set of reachable peers at time $t$, $\mathcal{N}_i(t)$, and selects one peer $j \in \mathcal{N}_i(t)$ according to the utility-guided policy detailed in Section~\ref{sec:peer_selection_policy}.
The selected peer provides $i$ with an additional \emph{gradient-based learning signal}: information about how its current $h_i$ should change to better fit data observed by peer $j$.

Specifically, node~$i$ sends its current head $h_i$ to peer $j$ (line~10).
Peer $j$ treats the received parameters as a temporary copy, leaving its own personalized head unchanged.
Using a local minibatch $\mathcal{B}_j$, it computes the gradient of its loss with respect to the received head (line~11).
This gradient identifies the parameter direction that would reduce the loss of $h_i$ on peer $j$'s observations, and is therefore returned to node~$i$ as the learning signal associated with that encounter.
This produces the \emph{remote gradient}:

\begin{equation}
    g_j = \nabla_h \ell_j(h;\mathcal{E}_j,\mathcal{B}_j),
\label{eq:pulse-remote-gradient}
\end{equation}

where $\ell_j(\cdot)$ is the loss of $j$ and $\mathcal{B}_j$ is a minibatch drawn from $D_j$.
In parallel, node~$i$ computes its \emph{local gradient} at the same parameter point (line~12),

\begin{equation}
    g_i = \nabla_h \ell_i(h;\mathcal{E}_i,\mathcal{B}_i),
\label{eq:pulse-local-gradient}
\end{equation}

using a local batch of data $\mathcal{B}_i$.

Gradient-based messages are a standard primitive in distributed optimization and collaborative learning~\cite{NIPS2017_f7552665}.
Here, however, the remote gradient has a specific role: since $g_j$ is evaluated at $h_i$, it describes how peer $j$'s data would modify node~$i$'s current personalized head, rather than how it would modify peer $j$'s own model.
PULSE can therefore assess whether this candidate gradient direction is compatible with node~$i$'s local learning objective before applying it.

To this end, we measure the cosine agreement between the two gradients (line~13):

\begin{equation}
    a_{ij} = \frac{\langle g_i,g_j\rangle} {\max\!\left(\lVert g_i\rVert\lVert g_j\rVert,\epsilon\right)},
\label{eq:pulse-agreement}
\end{equation}

where $\epsilon>0$ prevents numerical instability when either gradient has negligible norm.
The resulting score is approximately bounded in $[-1,1]$ and measures directional agreement independently of gradient magnitude.
A positive $a_{ij}$ indicates that the remote gradient $g_j$ has a component aligned with $i$'s local descent direction and is therefore expected to support the reduction of node~$i$'s loss.
Instead, a negative value identifies a conflicting direction, while $a_{ij}=0$ indicates that there is no benefit from the remote update.

This rationale is related to gradient-conflict mitigation methods in multi-task learning, which identifies incompatible gradients before combining them~\cite{NEURIPS2020_3fe78a8a}.
In PULSE, we use a lightweight alternative suited to brief encounters: rather than projecting or jointly optimizing conflicting gradients, it simply discards remote signals that do not agree with the local objective.
Moreover, even when $g_j$ is aligned with $g_i$, its norm may differ substantially because of local loss scales, batch composition, or data heterogeneity.
Without additional control, a large remote gradient could dominate the update despite providing only one peer's evidence.
We therefore rescale its magnitude to match that of the local gradient:

\begin{equation}
    \widetilde{g}_j =
    \sqrt{\frac{\lVert g_i\rVert^2}
    {\max\!\left(\lVert g_j\rVert^2,\epsilon\right)}}g_j.
\label{eq:pulse-normalized-gradient}
\end{equation}

The floor is therefore applied to the squared remote norm before taking the square root, matching the executable implementation.
When $\lVert g_j\rVert^2>\epsilon$, the resulting vector preserves the remote direction while matching its norm to that of $g_i$.
This provides a simple safeguard against magnitude mismatches between local and remote updates.
It is related in spirit to gradient-normalization approaches that account for learning signals with different magnitudes~\cite{pmlr-v80-chen18a}, although PULSE does not balance multiple task losses and uses normalization solely to scale the accepted remote gradient.

The node~$i$ also maintains a signed utility estimate
$\mathcal{U}_i[j]\in[-1,1]$ for each peer $j$.
This value summarizes whether, in previous encounters, the peer has tended to provide gradients aligned with or conflicting with node~$i$'s local objective.
The contribution of the remote signal (line~14) is set to

\begin{equation}
    \alpha_{ij} = \bar{\alpha} \max\!\bigl(\mathcal{U}_i[j],0\bigr) \max(a_{ij},0),
\label{eq:pulse-weight}
\end{equation}

where $\bar{\alpha}\in(0,1)$ is a fixed upper bound on the remote contribution.
Since $\alpha_{ij}\leq\bar{\alpha}<1$, the local gradient always retains a non-zero weight in the update.
Thus, a peer can contribute substantially only when it has been useful in previous encounters and its current gradient is strongly aligned with the local gradient.
Negative historical utility never increases the remote contribution; however, it remains relevant to the peer-selection policy, which may decide not to initiate communication with a peer that has repeatedly produced conflicting signals.

The local head of node~$i$ is then updated (line~15) as

\begin{equation}
    h_i \leftarrow h_i - \eta\left[(1-\alpha_{ij})g_i + \alpha_{ij}\widetilde{g}_j \right].
\label{eq:pulse-mixing}
\end{equation}

The update combines the local learning direction with a compatible, norm-matched remote direction.
Since $\alpha_{ij}\leq\bar{\alpha}$, choosing $\bar{\alpha}<1$ ensures that the local gradient always retains a non-zero contribution to the update.
When $a_{ij}>0$, the peer provides a direction consistent with the change suggested by node~$i$'s own data.
Otherwise, $\alpha_{ij}=0$ and~\eqref{eq:pulse-mixing} reduces to a standard local gradient step.

After a peer-assisted update, node~$i$ updates the utility of the selected peer (line~16) using the observed gradient agreement:

\begin{equation}
    \mathcal{U}_i[j] \leftarrow (1-\mu)\mathcal{U}_i[j] + \mu a_{ij},
\label{eq:pulse-utility-update}
\end{equation}

where $\mu\in(0,1]$ controls the weight assigned to the most recent interaction.
This recursion is an exponential moving average of the signed gradient agreement: positive values indicate that peer $j$ has generally supplied compatible learning directions, whereas negative values indicate recurrent conflict with node~$i$'s local objective.
The estimate therefore retains information from past encounters while adapting to changes in the peer's usefulness over time.

\subsection{Peer Selection Policy}
\label{sec:peer_selection_policy}

At each peer-discovery event, node~$i$ must decide whether communication is worthwhile and, if so, which currently reachable peer should be contacted.
This is a sequential decision problem: the usefulness of a peer can only be assessed after interacting with it, while its utility may vary over time as the users' contexts and data distributions evolve.

PULSE addresses this decision through a lightweight policy inspired by the \emph{Upper Confidence Bound} (UCB) principle~\cite{Auer2002}.
Accordingly, each reachable peer is treated as an alternative whose utility is learned online from the gradient agreements observed in previous encounters.
Crucially, PULSE also includes a \emph{no-contact} action, denoted by $\varnothing$, enabling node~$i$ to skip communication when none of the available peers appears sufficiently promising, rather than being forced to contact an arbitrary device (line 9).
Selecting $\varnothing$ means that node~$i$ does not engage in peer collaboration, while still taking advantage of the update opportunity through the local instance of Eq.~\eqref{eq:pulse-mixing}, with mixing coefficient equal to zero. This avoids unnecessary communication without interrupting local adaptation.

For every discovered peer $j$, node~$i$ maintains a utility estimate $\mathcal{U}_i[j]$ and a selection count $n_{ij}$.
The former summarizes whether the peer has historically provided aligned or conflicting gradients, whereas the latter measures the amount of evidence collected about that peer.
Let $s_i$ denote the number of peer-selection opportunities observed so far by node~$i$.
At a peer-discovery event, node~$i$ selects

\begin{equation}
j_i^\star
\gets
\operatorname*{arg\,max}_{j\in\mathcal{N}_i(t)\cup\{\varnothing\}}
\begin{cases}
\displaystyle
\mathcal{U}_i[j]
+
\beta
\sqrt{
    \frac{\log(2+s_i)}
    {1+n_{ij}}
},
& j\neq\varnothing, \\[10pt]
0,
& j=\varnothing,
\end{cases}
\label{eq:pulse-peer-selection}
\end{equation}

where $\beta>0$ controls the degree of exploration. The no-contact action has index zero, providing the reference against which the estimated benefit of a peer exchange is evaluated.

For a reachable peer $j$, the first term favors devices that have previously supplied gradients compatible with node~$i$'s local objective.
The second term is an \emph{exploration bonus}: it is larger for peers selected only a few times and decreases as $n_{ij}$ grows.
Consequently, a newly discovered peer is initially explored, whereas a peer that has repeatedly produced conflicting gradients may receive an index below that of the no-contact action.
In this case, PULSE avoids communication even if that peer is the only currently reachable device.
Moreover, the policy uses the encounter index $s_i$, rather than physical time, so that its exploration behavior evolves only when node~$i$ has an actual opportunity to make a communication decision.

After the decision, $s_i$ is incremented.
If $j_i^\star\neq\varnothing$, node~$i$ performs the peer-assisted update and increments $n_{ij_i^\star}$ after observing the resulting gradient agreement.
Over time, the policy identifies reachable peers that tend to provide beneficial learning signals, while avoiding exchanges whose expected contribution does not justify their cost.

\section{Simulation Results}

We firstly evaluate PULSE in a controlled event-driven simulator of decentralized learning among wearable devices under intermittent device-to-device connectivity. Each node represents one user and retains only its associated sensor data, while the simulator replays local-training opportunities and peer contacts over time.

We consider heterogeneous wearable applications and synthetic and real-world contact traces, which vary in sensing modality, learning task, user-data heterogeneity, and encounter dynamics. We compare PULSE with representative opportunistic-learning baselines in terms of classification performance and communication demand, measured by the number of completed peer-assisted message exchanges.

\subsection{Simulated Scenarios}
\label{sec:simulated_scenarios}

\begin{table}[t]
\centering
\caption{Summary of the evaluation datasets.}
\label{tab:datasets_summary}
\small
\begin{tabular}{llrrrr}
\toprule
Dataset & Scenario & Users & Samples & Labels (Max.\ \%) \\
\midrule
UCI-HAR & HAR & 30 & 10.32 K & 6 (18.9) \\
HHAR    & HAR & 9  & 87.6 K & 6 (19.5) \\
SisFall & FD  & 38 & 55.1 K & 2 (67.4) \\
WESAD   & SR  & 15 & 1 K  & 3 (54.5) \\
\bottomrule
\end{tabular}
\end{table}

We consider three wearable sensing applications: Human Activity Recognition (HAR), Fall Detection (FD), and Stress Recognition (SR).
As summarized in Table~\ref{tab:datasets_summary}, the scenarios differ in their sensing modalities, number of users, label spaces, and class imbalance.
\textbf{Human activity recognition (HAR).}
We adopt a cross-dataset setup based on UCI-HAR~\cite{uci_har} and HHAR~\cite{hhar}.
UCI-HAR comprises smartphone inertial recordings from
30 users performing six daily activities and is used solely to obtain a common pretrained initialization. Decentralized simulations are conducted on HHAR, which provides 87.6K accelerometer windows from nine users collected with heterogeneous smartphones and smartwatches.
This setting captures both user- and device-dependent distribution shifts between pretraining and deployment. For both datasets, we use three-axis accelerometer windows of 128 samples, independently normalized through per-axis $z$-score standardization.
\textbf{Fall detection (FD).}
SisFall~\cite{DBLP:journals/sensors/SucerquiaLB17} provides wearable accelerometer and gyroscope recordings from 38 users performing activities of daily living and falls. We formulate the task as binary fall versus non-fall classification.
As shown in Table~\ref{tab:datasets_summary}, its majority class accounts for 67.4\% of the samples, making it the most imbalanced scenario.
\textbf{Stress recognition (SR).}
We use WESAD~\cite{10.1145/3242969.3242985}, which contains synchronized wrist- and chest-worn sensor measurements from 15 users exposed to different affective conditions.
We consider its three-class stress-recognition task.
Unlike the inertial scenarios, its heterogeneity mainly stems from subject-specific physiological responses.

For FD and SR, we partition the available subjects into disjoint
\emph{pretraining}, \emph{validation}, and \emph{simulation} sets.
Pretraining data are used to learn a task-specific initial encoder, validation data are used exclusively for checkpoint selection, and simulation data remain unseen until the controlled experiments.
In FD, we use 22 subjects (29,187 samples, 53.0\%) for pretraining, 6 subjects (9,832 samples, 17.8\%) for validation, and 10 subjects (16,070 samples, 29.2\%) for the decentralized simulation.
The corresponding SR partitions contain 3 subjects and 204 samples (19.9\%), 2 subjects and 140 samples (13.6\%), and 10 subjects and 682 samples (66.5\%), respectively.

HAR follows a different cross-dataset protocol.
UCI-HAR contributes 26 subjects and 8,921 samples (86.6\%) to pretraining, and four subjects and 1,378 samples (13.4\%) to validation.
All nine HHAR users and their 87,634 samples are instead reserved for the decentralized simulation.
Consequently, the HAR setting evaluates adaptation not only to previously unseen users, but
also to a distinct target dataset collected with heterogeneous smartphones and smartwatches.

\subsection{Models Pretraining}
\label{sec:models_pretraining}

\begin{figure}[t]
    \centering
    \begin{subfigure}[b]{0.48\columnwidth}
        \centering
        \includegraphics[width=\linewidth]{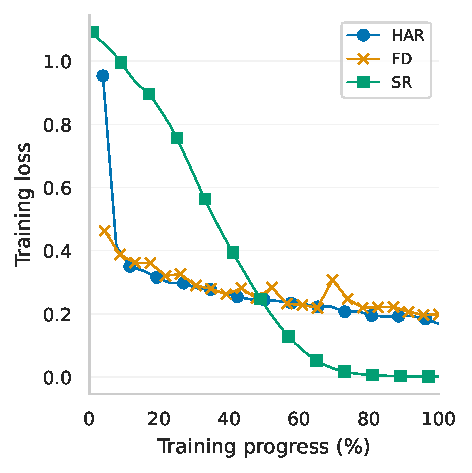}
        \caption{Training loss.}
        \label{fig:pretraining_loss}
    \end{subfigure}
    \hfill
    \begin{subfigure}[b]{0.48\columnwidth}
        \centering
        \includegraphics[width=\linewidth]{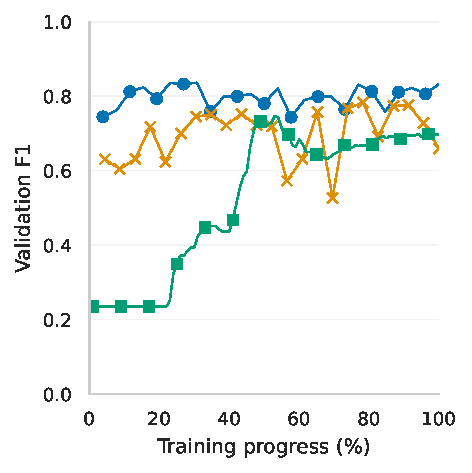}
        \caption{Validation F1-score.}
        \label{fig:pretraining_validation_f1}
    \end{subfigure}
    \caption{Pretraining curves for the task-specific models.}
    \label{fig:pretraining_curves}
\end{figure}

To provide all nodes with a common yet task-aware initialization, we pretrain a
lightweight encoder for each scenario before the decentralized simulations.
The architectures are tailored to the available sensing modalities while
maintaining a common 64-dimensional embedding space and a limited computational
footprint. During pretraining, each encoder is paired with an auxiliary MLP
classifier comprising one 64-unit hidden layer. The selected encoder checkpoint
then initializes every simulated node, which locally fine-tunes it together
with its own personalized classifier.

For HAR, we use a compact one-dimensional CNN with two temporal convolutional
blocks over the three-axis accelerometer signal. FD uses a six-channel temporal
convolutional network with residual dilated blocks to process accelerometer and
gyroscope windows. For SR, separate convolutional branches encode the
accelerometer, blood-volume pulse, electrodermal activity, and temperature
signals; their representations are fused before classification.
All models are optimized with AdamW and cross-entropy loss using a learning
rate of $10^{-3}$. We monitor validation F1-score, retaining the checkpoint with
the highest value, and restore it before deployment. Early stopping uses a
minimum improvement of $0.001$: HAR and FD stop after 20 and 5 epochs without
improvement, respectively, whereas SR is trained for the maximum budget of
100 epochs with a patience of 50.

Figure~\ref{fig:pretraining_curves} shows the pre-training dynamics, with the
horizontal axis normalized by the training budget of each scenario. The
training loss decreases consistently in all cases. The selected HAR, FD, and
SR encoders achieve validation F1-scores of 0.836, 0.783, and 0.746 at epochs
8, 18, and 53, respectively. Validation F1 is not strictly monotonic,
particularly for FD and SR, as expected under stochastic optimization and
limited validation data.
Selecting the best validation checkpoint therefore provides a robust initialization before local personalization in the opportunistic scenarios.

\subsection{Contact Traces}
\label{sec:contact_traces}

\begin{figure}[t]
    \centering
    \includegraphics[width=0.89\columnwidth]{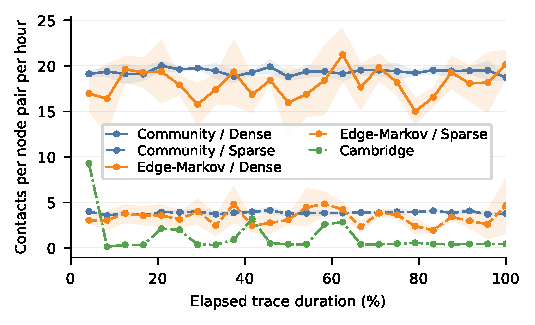}
    \caption{Temporal evolution of contact opportunities.}
    \label{fig:contact_temporal_activity}
\end{figure}

We replay time-stamped pairwise contact traces to determine when devices can exchange learning signals. A contact event indicates that the corresponding node pair can communicate at a given time. Repeated observations of the same undirected pair within a snapshot are removed. We consider two controlled synthetic mobility models and the real-world Cambridge/Haggle Bluetooth trace~\cite{c70011-22}.

\subsubsection{Synthetic traces}
We generate traces using the \emph{Community}~\cite{DBLP:conf/mobihoc/MusolesiM06} and \emph{Edge-Markov}~\cite{DBLP:conf/podc/ClementiMMPS08} models. Each synthetic trace contains ten nodes, spans 24 hours, and is sampled every 20 seconds. For each model, we generate three independent realizations using seeds 7, 42, and 123.
For both models, we consider a \emph{Dense} and a \emph{Sparse} connectivity regime that differ in the frequency of communication opportunities while preserving the defining property of the respective mobility model.
Dense traces provide approximately 4.5--4.9 active contacts per snapshot, corresponding to a mean occupancy of about 10\% of all possible node pairs. Sparse traces provide approximately 0.8--1.0 contacts per snapshot, or about 2\% occupancy. They therefore reduce the rate of peer encounters by roughly a factor of five, creating a more constrained setting for opportunistic learning.

In Community traces, contacts remain predominantly local: approximately 86.6\% and 86.3\% of events occur within communities in the Dense and Sparse regimes, respectively.
For each mobility model, the two regimes retain the same node population and contact-generation mechanism, while varying only the overall frequency of contact opportunities. The Dense--Sparse comparison therefore isolates the effect of peer availability under the same underlying mobility model.

\subsubsection{Real trace}
We additionally use the Cambridge/Haggle trace, which records Bluetooth encounters among iMotes carried by members of a research group at the University of Cambridge~\cite{c70011-22}. It contains 12 nodes, spans approximately 126.6 hours, and is sampled every 120 seconds. Its mean pair occupancy is 4.14\%, intermediate between the controlled Dense and Sparse regimes. Unlike the approximately stationary synthetic traces, however, Cambridge exhibits temporally uneven activity, with contacts concentrated in a limited number of intervals. Figure~\ref{fig:contact_temporal_activity} highlights both the controlled separation between Dense and Sparse conditions and the irregular temporal dynamics of the real trace. Together, these traces assess the methods under frequent, scarce, and naturally bursty peer-learning opportunities.

\begin{table*}[t]
\centering
\caption{Final test performance. Bold and underlined values indicate the best
and second-best available methods, respectively, for each dataset,
contact-trace configuration, and metric.}
\label{tab:final_results}
\small
\renewcommand{\arraystretch}{1.10}
\resizebox{\textwidth}{!}{%
\begin{tabular}{ll*{15}{r}}
\toprule
& & \multicolumn{3}{c}{Cambridge}
& \multicolumn{3}{c}{Community--Dense}
& \multicolumn{3}{c}{Community--Sparse}
& \multicolumn{3}{c}{Edge-Markov--Dense}
& \multicolumn{3}{c}{Edge-Markov--Sparse} \\
\cmidrule(lr){3-5}
\cmidrule(lr){6-8}
\cmidrule(lr){9-11}
\cmidrule(lr){12-14}
\cmidrule(l){15-17}
Scenario & Method
& Loss & Acc. & F1
& Loss & Acc. & F1
& Loss & Acc. & F1
& Loss & Acc. & F1
& Loss & Acc. & F1 \\
\midrule
\multirow{5}{*}{HAR}
& Local
& \underline{.40} & \textbf{.87} & \underline{.86}
& .88 & .68 & .64
& \underline{.89} & \underline{.68} & \underline{.64}
& .87 & .68 & .64
& \underline{.88} & \underline{.68} & \underline{.64} \\
& OppAVG
& .41 & \underline{.86} & \underline{.86}
& .96 & .61 & .56
& .97 & .61 & .56
& .96 & .61 & .55
& .97 & .61 & .55 \\
& OppFL
& 1.05 & .55 & .52
& .91 & .65 & .64
& 1.24 & .54 & .51
& .94 & .65 & .64
& 1.20 & .54 & .51 \\
& OPTIMIST
& .91 & .59 & .57
& \underline{.86} & \underline{.70} & \underline{.69}
& .95 & .65 & .63
& \underline{.83} & \underline{.71} & \underline{.70}
& 1.00 & .65 & .63 \\
& PULSE
& \textbf{.39} & \textbf{.87} & \textbf{.87}
& \textbf{.55} & \textbf{.82} & \textbf{.82}
& \textbf{.71} & \textbf{.77} & \textbf{.76}
& \textbf{.58} & \textbf{.82} & \textbf{.81}
& \textbf{.69} & \textbf{.76} & \textbf{.75} \\
\midrule
\multirow{5}{*}{FD}
& Local
& \underline{.09} & \textbf{.92} & \textbf{.82}
& \underline{.19} & \textbf{.91} & \textbf{.81}
& \textbf{.19} & \textbf{.91} & \textbf{.81}
& \underline{.19} & \textbf{.91} & \textbf{.81}
& \textbf{.19} & \textbf{.91} & \textbf{.81} \\
& OppAVG
& \underline{.09} & \textbf{.92} & \textbf{.82}
& \underline{.19} & \textbf{.91} & \textbf{.81}
& \textbf{.19} & \textbf{.91} & \textbf{.81}
& \underline{.19} & \textbf{.91} & \textbf{.81}
& \textbf{.19} & \textbf{.91} & \textbf{.81} \\
& OppFL
& .20 & \underline{.85} & \underline{.74}
& .30 & \underline{.86} & \underline{.76}
& .35 & \underline{.84} & .74
& .29 & \underline{.86} & \underline{.76}
& .28 & .84 & .74 \\
& OPTIMIST
& .31 & .84 & \underline{.74}
& .41 & .84 & .75
& .40 & \underline{.84} & .74
& .39 & .85 & .75
& .36 & \underline{.85} & \underline{.75} \\
& PULSE
& \textbf{.08} & \textbf{.92} & \textbf{.82}
& \textbf{.18} & \textbf{.91} & \textbf{.81}
& \underline{.21} & \textbf{.91} & \underline{.80}
& \textbf{.18} & \textbf{.91} & \textbf{.81}
& \underline{.20} & \textbf{.91} & \textbf{.81} \\
\midrule
\multirow{5}{*}{SR}
& Local
& \textbf{.00} & \textbf{.98} & \textbf{.92}
& .10 & \textbf{.95} & \underline{.87}
& .11 & \underline{.94} & \underline{.86}
& .10 & \textbf{.95} & \underline{.87}
& .10 & \textbf{.95} & \underline{.87} \\
& OppAVG
& \textbf{.00} & \underline{.97} & \textbf{.92}
& \underline{.09} & \underline{.93} & .85
& \underline{.10} & .93 & .85
& \underline{.09} & \underline{.93} & .84
& \underline{.09} & \underline{.94} & .85 \\
& OppFL
& .29 & .84 & .69
& .21 & .86 & .75
& .45 & .83 & .66
& .25 & .87 & .75
& .45 & .83 & .66 \\
& OPTIMIST
& .16 & .86 & .77
& .20 & .88 & .77
& .34 & .86 & .76
& .18 & .88 & .77
& .29 & .86 & .75 \\
& PULSE
& \underline{.01} & \underline{.97} & \underline{.91}
& \textbf{.04} & \textbf{.95} & \textbf{.88}
& \textbf{.06} & \textbf{.95} & \textbf{.87}
& \textbf{.04} & \textbf{.95} & \textbf{.88}
& \textbf{.05} & \textbf{.95} & \textbf{.88} \\
\bottomrule
\end{tabular}%
}
\end{table*}

\subsection{Baselines}
\label{sec:baselines}

We compare PULSE with four baselines.
\emph{Local} trains exclusively on private data, without any communications.
\emph{OppAVG} opportunistically averages the models of all reachable peers.
\emph{OppFL} sends the local model to a peer, receives a gradient computed on the peer's private minibatch, and aggregates it with the local gradient.
\emph{OPTIMIST} partitions the classifier into \(N\) subnetworks, one initially assigned to each of the \(N\) nodes; mutually matched nodes exchange their active subnetworks and update only the received one.
It performs no update without an accepted exchange.

All methods start from the same pretrained model, update both the encoder and prediction head, and never exchange raw data.
They use two SGD steps, batch size of $16$, and learning rate of $0.01$.
OppFL uses a zero similarity threshold, unit peer-gradient weight, two interaction rounds, and momentum aggregation, thus allowing the method to exploit all non-conflicting peer updates while preserving its original aggregation mechanism.
For OPTIMIST, we set the number of subnetworks as to the number of nodes in each simulation. For PULSE, we set the maximum mixing coefficient to $\bar{\alpha}=0.5$, the utility-update rate to $\mu=0.2$, and the UCB exploration coefficient to $\beta=0.25$. These values keep peer contributions bounded, let utility estimates react to recent interactions without becoming overly volatile, and provide limited exploration of previously untested peers.

\subsection{Classification Results}
\label{sec:classification_results}

Table~\ref{tab:final_results} reports the final performance across the three real-world sensing applications and five contact-trace configurations. For each simulation user, local data are split into 70\% for training and 30\% for testing, with Accuracy and Macro-F1 evaluated exclusively on the local test set. Local learning is already a strong baseline in several settings: each node is therefore trained and evaluated on data from the same user, all nodes start from a pretrained encoder, and the simulated users generally retain examples from every target class. Hence, collaboration is beneficial only when encountered peers provide complementary and compatible learning signals.

This effect is particularly evident in FD.
Local, OppAVG, and PULSE reach the same performance with the Cambridge trace (0.92 Accuracy and 0.82 F1), and remain closely aligned in the dense synthetic scenarios (approximately 0.91 Accuracy and 0.81 F1).
Sparse contacts lead only to a limited reduction for PULSE, whose F1 reaches 0.80 under Community--Sparse. Since FD is a binary task and its encoder is pretrained on data from other subjects of the same dataset, the strong Local results suggest limited headroom for collaboration.
PULSE nevertheless remains competitive, showing that its peer-assisted updates do not introduce substantial negative transfer when local personalization is already effective.

A similar pattern emerges in SR.
Local obtains the best Cambridge accuracy (0.98), while PULSE remains close at 0.97 accuracy and 0.91 F1.
Under all synthetic traces, however, PULSE achieves the highest F1, reaching 0.88 in the dense configurations and 0.87--0.88 under sparse connectivity, while maintaining 0.95 accuracy throughout. Although the margins over Local are moderate, these results show that PULSE can exploit cross-user information without compromising personalized performance.

The benefits of collaboration are substantially clearer in HAR, where the encoder is pretrained on UCI-HAR and then adapted to users and devices from a different dataset.
In the dense scenarios, PULSE reaches 0.82 accuracy and 0.81--0.82 F1, compared with 0.68 accuracy and 0.64 F1 for Local.
OPTIMIST is the strongest baseline in these settings, but remains below PULSE (0.70--0.71 accuracy and 0.69--0.70 F1).
PULSE also retains a clear advantage under sparse contacts, obtaining 0.76--0.77 accuracy and 0.75--0.76 F1.
In contrast, OPTIMIST and OppFL fall to approximately 0.65/0.63 and 0.54/0.51, respectively, reflecting their stronger dependence on frequent successful encounters.
With Cambridge, PULSE matches Local in accuracy (0.87) and slightly improves F1 (0.87 versus 0.86), whereas
OppFL and OPTIMIST reach 0.55/0.52 and 0.59/0.57, respectively.
This is consistent with the intermittent nature of the real trace, where long periods with few contacts penalize approaches whose optimization is primarily encounter-driven.

Overall, PULSE provides the most consistent behavior across the evaluation.
It achieves the best accuracy and F1 in all HAR configurations, the highest Macro-F1 in every synthetic SR scenario, and remains comparable to Local on FD.
The final-loss values follow the same trend, with PULSE attaining the lowest rounded loss in all HAR settings and in most synthetic FD and SR settings.
These results indicate that selective collaboration can preserve strong local personalization when communication is unhelpful, while providing substantial gains when local data and pretrained representations are less well aligned with the target setting.

\subsection{Communication Efficiency}
\label{sec:sim_contacts}
\begin{table}[t]
    \centering
    \caption{Average number ($\times 1000$) of completed peer-assisted updates across the
    simulation.}
    %C: Cambridge; CD/CS: Community
    %Dense/Sparse; EMD/EMS: Edge-Markov Dense/Sparse.}
    \label{tab:completed_contacts}
    \small
    \begin{tabular}{lrrrr}
        \toprule
        \textbf{Scenario} &
        \textbf{OppAVG} &
        \textbf{OppFL} &
        \textbf{OPTIMIST} &
        \textbf{PULSE} \\
        \midrule
        HAR--C   & 10.99 & 10.99 & 5.31  & \textbf{1.42} \\
        HAR--CD  & 33.35 & 33.35 & 18.52 & \textbf{3.06} \\
        HAR--CS  & 6.66  & 6.66  & 5.90  & \textbf{1.87} \\
        HAR--EMD & 31.20 & 31.20 & 15.86 & \textbf{3.13} \\
        HAR--EMS & 5.79  & 5.79  & 5.04  & \textbf{2.42} \\
        \midrule
        FD--C   & 14.81 & 11.93 & 7.35  & \textbf{1.50} \\
        FD--CD  & 41.80 & 33.30 & 21.36 & \textbf{3.44} \\
        FD--CS  & 8.36  & 6.64  & 7.28  & \textbf{1.97} \\
        FD--EMD & 38.94 & 27.65 & 18.05 & \textbf{3.54} \\
        FD--EMS & 7.28  & 5.82  & 6.22  & \textbf{2.30} \\
        \midrule
        SR--C   & 14.81 & 14.81 & 7.35  & \textbf{1.66} \\
        SR--CD  & 41.80 & 41.80 & 21.36 & \textbf{2.68} \\
        SR--CS  & 8.36  & 8.36  & 7.28  & \textbf{1.99} \\
        SR--EMD & 38.94 & 38.94 & 18.05 & \textbf{2.97} \\
        SR--EMS & 7.28  & 7.28  & 6.22  & \textbf{2.43} \\
        \bottomrule
    \end{tabular}
\end{table}

Table~\ref{tab:completed_contacts} reports the average number ($\times 1000$) of peer-assisted updates completed throughout the simulated scenarios.
As expected, dense synthetic traces generate substantially more exploitable interactions than their sparse counterparts, while the Cambridge trace typically lies between the two regimes.
OppAVG performs the largest number of updates, ranging from 5,788 to 41,801 across the considered settings.
OppFL exhibits a similar communication demand in HAR and SR, whereas in FD it completes between 5,817 and 33,301 updates due to its similarity-based peer filtering.
OPTIMIST reduces the number of interactions relative to these methods, but still performs between 5,044 and 21,358 updates.

In contrast, PULSE completes only 1,420--3,541 peer-assisted updates across all scenarios, corresponding to an approximately 80--94\% reduction with respect to OppAVG.
This substantially lower message-exchange demand is a practical advantage for the resource-constrained wearable devices considered in this work, where repeated communication incurs energy and latency costs.
Importantly, as shown by the performance results discussed next, this reduction
does not come at the expense of personalization effectiveness.

\section{On-Device Evaluation}
\label{sec:on_device_feasibility}

\begin{figure}[t]
    \centering
    \includegraphics[width=\linewidth]{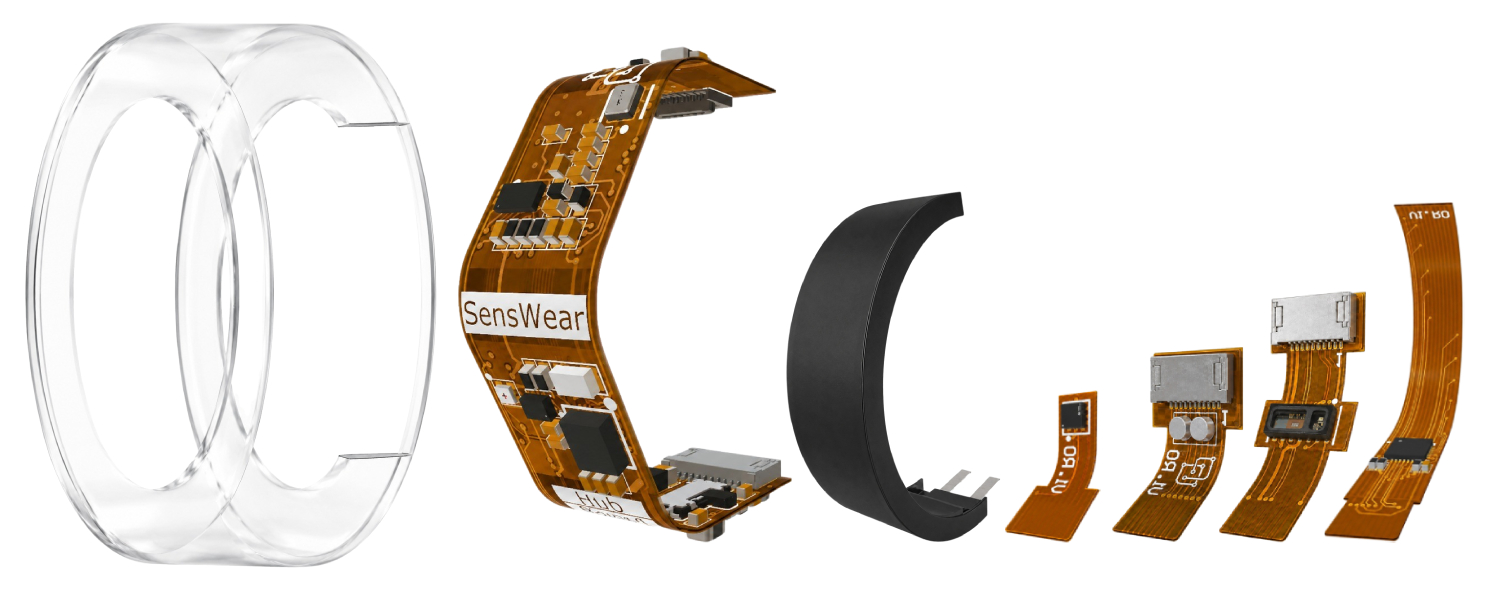}
    \caption{Exploded view of the SensWear ring configuration: enclosure, flexible main board, curved battery, and interchangeable daughter boards.}
    %The PULSE prototype uses the main-board nRF54L15 processor and Bluetooth Low
    %Energy radio, BHI360 IMU, local non-volatile storage, and power subsystem.
    %The optional sensing and actuation daughter boards at right are shown for context and are not required by PULSE.}
    \label{fig:senswear_platform}
\end{figure}

We evaluate PULSE's feasibility on SensWear, an open-source modular wearable platform based on Zephyr RTOS~\cite{senswear_github, zephyr}.
Its flexible main board (Fig.~\ref{fig:senswear_platform}) integrates a Nordic nRF54L15 wireless SoC with a 128-MHz Arm Cortex-M33 processor, Bluetooth Low Energy (BLE) connectivity, 256~KB of RAM, and 1.5~MB of non-volatile memory~\cite{nrf54l15_web}. A BHI360 six-axis IMU provides inertial sensing, while 4~MiB of external non-volatile memory supports local storage of sensor samples and model parameters.

At the firmware level, PULSE learning operations are mapped to event-driven Zephyr work items. During an opportunistic exchange, the device requesting peer assistance acts as the \emph{initiator} (node $i$), while the selected peer acts as the \emph{responder} (node $j$).
The initiator sends its personalized head $h_i$ to the responder, which evaluates it on a private local minibatch and returns the resulting gradient $g_j$. The initiator then incorporates the received gradient into its local update, completing the peer-assisted interaction without exchanging raw sensor data.

%The firmware maps each PULSE event to a Zephyr work item. The pretrained encoder $\mathcal{E}$ is read-only, whereas the personalized head $h_i$, gradients, and peer utilities are mutable. A contacted initiator serializes only $h_i$; the responder evaluates the received head on a private minibatch and returns only $g_j$, so raw observations remain local. For the 64-dimensional embedding and the stated MLP head with one 64-unit hidden layer, the number of trainable parameters is
%\begin{equation}
%    P(C)=64\!\times\!64+64+64C+C=4160+65C,
%    \label{eq:head_parameter_count}
%\end{equation}
%where $C$ is the number of output classes. With $q$ bytes per serialized parameter, one successful request--response transfers $2qP(C)$ application bytes plus message headers. Link-layer traffic is not inferred from application payload size.

\subsection{Experimental Protocol}

%We compile a release benchmark image and use the same model structure, batch size of 16, and $R=2$ local steps used in the controlled evaluation. The available current capture used the same model artifact. During timing runs, structured UART records are emitted only after the measured interval of each event. The shell and visual indicators are disabled, and any instrumentation retained is identical in all compared configurations.

%The timing paths are run for five warm-up repetitions followed by 100 measured trials on one physical board pair. The completed initiator current capture has the same counts for local-update and scheduled connected-path windows. We report the mean and median together with the interquartile range and p5--p95 whiskers because radio transactions have skewed tail latency. An otherwise identical sensing-and-BLE image without PULSE provides the baseline for incremental flash and static RAM costs. Event current is compared with a pre-event baseline from the same current stream.

All measurements use the same local model and hyperparameters adopted in the HAR simulations. Timing and energy are evaluated on two communicating SensWear devices, with five warm-up repetitions followed by 100 measured trials for each execution path, while instrumentation is kept outside the measured intervals to avoid affecting the results. Latency is characterized through the mean, median, interquartile range, and p5--p95 intervals to capture variability in wireless communication, while energy is derived from event-level current measurements relative to the corresponding pre-event baseline. Incremental flash and static RAM requirements are measured against an otherwise identical sensing-and-BLE firmware image without PULSE, isolating the memory overhead introduced by the framework.

\begin{figure*}[t]
    \centering
    \begin{subfigure}[t]{0.40\textwidth}
        \centering
        \begin{minipage}[c][0.725\linewidth][c]{\linewidth}
            \centering
            \includegraphics[width=\linewidth,height=0.725\linewidth,
            keepaspectratio]{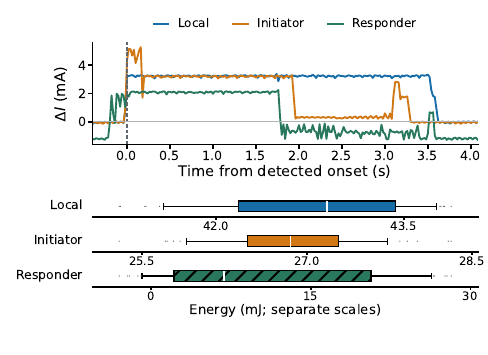}
        \end{minipage}
        \phantomsubcaption\label{fig:ondevice_power}
        \par\smallskip{\small(a)}
    \end{subfigure}
    \hfill
    \begin{subfigure}[t]{0.29\textwidth}
        \centering
        \begin{minipage}[c][\linewidth][c]{\linewidth}
            \centering
            \includegraphics[width=\linewidth,height=0.995\linewidth,
            keepaspectratio]{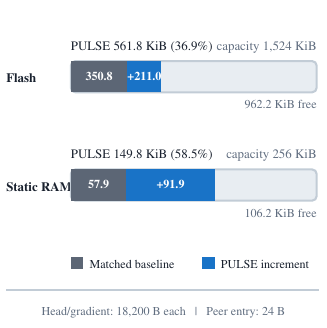}
        \end{minipage}
        \phantomsubcaption\label{fig:ondevice_resources}
        \par\smallskip{\small(b)}
    \end{subfigure}
    \hfill
    \begin{subfigure}[t]{0.29\textwidth}
        \centering
        \begin{minipage}[c][\linewidth][c]{\linewidth}
            \centering
            \includegraphics[width=\linewidth,height=0.995\linewidth,
            keepaspectratio]{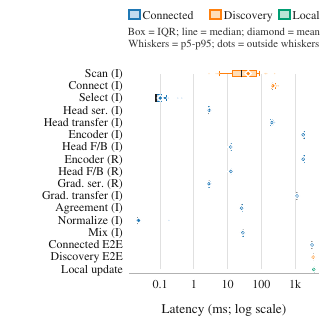}
        \end{minipage}
        \phantomsubcaption\label{fig:ondevice_latency}
        \par\smallskip{\small(c)}
    \end{subfigure}
    %\caption{On-device feasibility of PULSE. (a) Aaligned current traces and baseline-subtracted energy distributions for local, scheduled connected-path initiator, and responder-candidate windows. Each distribution contains 100 windows after five warm-ups. The lower boxplots use separate labeled mJ scales to reveal within-role spread. The roles were recorded separately, so these are not paired exchange energies. (b) Static footprint from matched final-build linker maps. Each bar decomposes the PULSE image into the sensing-and-BLE baseline and incremental PULSE footprint; the remainder is unused capacity. Values are KiB (1,024 bytes). (c) Latency distributions on one SensWear pair. Diamonds show means. IQR boxes, median lines, and p5--p95 whiskers appear only when distinguishable from the diamond at print size; separated outliers remain as dots. Shared stages show the connected path, while discovery adds scan, connection, and end-to-end latency. End-to-end $n=100$; stage $n=96$--100.}
    \caption{On-device feasibility of PULSE. (a) Energy distributions for local, initiator, and responder operations; (b) static memory footprint of the PULSE implementation; and (c) latency distributions measured on a SensWear device pair. Energy roles were measured separately and therefore do not represent paired exchanges.}
    \label{fig:ondevice_results}
\end{figure*}

\subsection{Power and Energy}

Power consumption is measured using a Keysight 34465A digital multimeter connected in series with a current-limited 3.7-V supply at the SensWear battery input, while the peer device is powered independently. Current is sampled at 50~Hz, with the initiator and responder roles profiled in separate runs.
For each event, baseline power is estimated over the 0.4~s preceding its detected onset, and incremental energy is obtained by integrating the power above this baseline over the following 4.1~s, including the return to idle. The resulting value represents source-side energy measured at the board input, while the responder energy is estimated assuming the same 3.7-V supply voltage.

As shown in Fig.~\ref{fig:ondevice_power}, the energy required by PULSE remains particularly limited across all profiled operations. Local adaptation is the most demanding operation, with a median incremental energy of 42.88~mJ (IQR 42.18--43.43~mJ), whereas initiating an opportunistic exchange requires only 26.85~mJ (IQR 26.46--27.28~mJ).
The responder incurs a considerably smaller median cost of 6.87~mJ (IQR 2.15--20.65~mJ), indicating that assisting a nearby device introduces only a modest additional energy demand.
These results show that peer-assisted learning does not dominate the energy cost of on-device adaptation, despite involving both local computation and wireless communication. This behavior is also reflected in the current traces, where local adaptation sustains elevated consumption for a longer interval, while peer collaboration concentrates its energy demand in shorter computation and communication phases.
The measured energy profile therefore supports the feasibility of selective peer assistance on battery-powered wearable devices, particularly when exchanges are performed only when considered beneficial by PULSE.

\subsection{Memory Usage}

%Static flash and RAM are obtained from clean, matched final-capture builds of the release and sensing-and-BLE baseline at the same firmware revision. We use the linker-defined usable regions as denominators and subtract the baseline totals from the PULSE totals to isolate PULSE's incremental footprint. The reported RAM is statically reserved memory from the linker map.

%Figure~\ref{fig:ondevice_resources} shows that the PULSE image occupies 561.8~KiB of the 1,524-KiB flash region (36.9\%), compared with 350.8~KiB for the matched baseline. PULSE therefore adds 211.0~KiB of flash and leaves 962.2~KiB free. Static RAM increases from 57.9 to 149.8~KiB, an incremental 91.9~KiB, leaving 106.2~KiB of the 256-KiB region free (58.5\% used). The implemented six-class FP32 head contains 4,550 parameters, so a serialized head or gradient occupies 18,200~bytes. Compiler-reported peer bookkeeping requires 24~bytes per peer; with the configured eight-peer limit, the complete utility table occupies 208~bytes including table-level state.

Flash and static RAM requirements are obtained from the linker maps of the PULSE firmware and an otherwise identical sensing-and-BLE baseline built from the same firmware revision. The difference between the two configurations is used to isolate the incremental memory footprint introduced by PULSE, with the linker-defined usable regions taken as the available device capacity.

As shown in Fig.~\ref{fig:ondevice_resources}, the complete PULSE firmware occupies 561.8~KiB of flash (36.9\% of the available 1,524~KiB) and 149.8~KiB of static RAM (58.5\% of 256~KiB), leaving 962.2~KiB and 106.2~KiB available, respectively.
Compared with the considered baseline, PULSE introduces an additional 211.0~KiB of flash and 91.9~KiB of RAM, showing that the complete learning framework fits within the memory resources of the wearable while retaining substantial capacity for other application functionality.
The classification head contains 4,550 parameters and occupies 18.2~KB when serialized, with the same size required for an exchanged gradient.
In contrast, the additional state required for peer selection is negligible, with the complete utility table for the configured eight peers requiring only 208~bytes.
The remaining memory capacity confirms that PULSE can be practically deployed alongside the sensing and communication functionality of a resource-constrained wearable.

\subsection{Execution Time}

%The firmware's 1-MHz wall counter measures each stage and the complete event without including subsequent UART export. Already-connected trials isolate the PULSE protocol, whereas discovery-to-update trials include link establishment.
%Figure~\ref{fig:ondevice_latency} shows mean diamonds for every phase and IQR boxes, median lines, and p5--p95 whiskers only where their printed spans remain distinguishable from the diamond. Separated outliers remain visible as dots; omitted marks do not indicate missing measurements. Shared stages show the connected path once; discovery adds scan, connection, and end-to-end. All end-to-end records are complete ($n=100$); malformed UART stage records are excluded without imputation ($n=96$--100).

%The local update completes in 3539.6~ms at the median and 3540.5~ms at p95. An already-connected accepted exchange takes 3167.5/3220.2~ms at the median/p95, while discovery-to-update takes 3428.4/3522.6~ms. Thus, discovery adds 260.9~ms (8.2\%) to the median transaction; the 26.2-ms scan and 210.3-ms connection medians explain most of this increase. Encoder execution (1740.0~ms), returned-gradient transfer (1132.5~ms), and head transfer (200.0~ms) dominate the connected path. Peer selection and normalization take only 0.075 and 0.021~ms, while agreement and mixing take 25.9 and 28.1~ms. All 100 initiator trials succeed in both accepted paths, giving a 0\% measured failure rate. The larger discovery p95-minus-median gap (94.2 versus 52.6~ms) localizes most tail variability to link establishment.

Execution time is measured using the firmware's 1-MHz wall counter, excluding subsequent logging operations. We profile both already-connected exchanges, which isolate the PULSE protocol, and discovery-to-update transactions that additionally include BLE link establishment. Each end-to-end path is evaluated over 100 trials, with latency distributions reported in Fig.~\ref{fig:ondevice_latency}.

Local adaptation completes in 3539.6~ms at the median (3540.5~ms at p95), while an already-connected opportunistic exchange requires 3167.5~ms (3220.2~ms at p95). Including peer discovery and connection increases the median latency to 3428.4~ms, corresponding to an overhead of only 260.9~ms (8.2\%) over the connected path.
Most of this overhead is attributable to scanning and connection establishment, while peer selection and gradient processing introduce comparatively small computational costs. All 100 initiator trials successfully complete in both exchange configurations, showing that the complete PULSE interaction can be reliably executed directly on a wearable device within just a few seconds.

\section{Conclusions and Future Work}
\label{sec:conclusion}

This paper presented PULSE, an event-driven framework for Personalized Opportunistic Learning on resource-constrained wearable devices. PULSE treats transient encounters as selective collaboration opportunities, combining utility-guided peer selection with gradient-level compatibility assessment. Across heterogeneous wearable applications and mobility conditions, PULSE consistently preserves or improves personalized performance while requiring substantially fewer peer-assisted exchanges than less selective opportunistic approaches. These results show that effective collaboration does not require exploiting every available encounter, highlighting the importance of selectivity in opportunistic settings.
Moreover, the prototype implementation further demonstrates the feasibility of the proposed mechanisms on wearable-class hardware.

Future work will extend PULSE towards more adaptive forms of opportunistic collaboration, where devices can dynamically select the knowledge to exchange and the most appropriate strategy for integrating it according to their current context, learning needs, and resource constraints.
We also plan to investigate richer context-aware peer-utility models that explicitly account also for communication and energy costs, and to evaluate long-term personalized adaptation in larger-scale real-world deployments under naturally occurring mobility patterns.

%\section*{Acknowledgment}

\bibliographystyle{IEEEtran}
\bibliography{references}
\end{document}